# AI-Facilitated Analysis of Abstracts and Conclusions: Flagging Unsubstantiated Claims and Ambiguous Pronouns

Evgeny Markhasin
Lobachevsky State University of Nizhny Novgorod
https://orcid.org/0000-0002-7419-3605
https://linkedin.com/in/evgenymarkhasin

**Abstract**

We present and evaluate a suite of proof-of-concept (PoC), structured workflow prompts designed to elicit human-like hierarchical reasoning while guiding Large Language Models (LLMs) in the high-level semantic and linguistic analysis of scholarly manuscripts. The prompts target two non-trivial analytical tasks within academic summaries (abstracts and conclusions): identifying unsubstantiated claims (informational integrity) and flagging semantically confusing ambiguous pronoun references (linguistic clarity). We conducted a systematic, multi-run evaluation on two frontier models (Gemini Pro 2.5 Pro and ChatGPT Plus o3) under varied context conditions. Our results for the informational integrity task reveal a significant divergence in model performance: while both models successfully identified an unsubstantiated head of a noun phrase (95% success), ChatGPT consistently failed (0% success) to identify an unsubstantiated adjectival modifier that Gemini correctly flagged (95% success), raising a question regarding the potential influence of the target's syntactic role. For the linguistic analysis task, both models performed well (80-90% success) with full manuscript context. Surprisingly, in a summary-only setting, Gemini's performance was substantially degraded, while ChatGPT achieved a perfect (100%) success rate. Our findings suggest that while structured prompting is a viable methodology for complex textual analysis, prompt performance may be highly dependent on the interplay between the model, task type, and context, highlighting the need for rigorous, model-specific testing.

**Keywords:** AI-assisted, AI-powered, AI-enhanced, automated, machine learning, academic summary.

## 1. Introduction

Computer-assisted tools for academic writing have a long history [1], but the recent emergence of state-of-the-art (SOTA) large language models (LLMs) has enabled new forms of accessible semantic and linguistic analysis [2, 3] and synthesis [4]. Modern LLM architectures, such as Gemini Pro 2.5 Pro [5], ChatGPT Plus o3 [6], and Claude Opus 4 [7], possess capabilities that are particularly useful for the semantic and structural analysis of technical texts. Specifically, their large context windows allow for the analysis of full-length research papers [8, 9], while in-context learning (ICL) [10–12] combined with advanced prompting strategies [13–21] makes it possible to systematically focus the model's attention on specialized aspects of a manuscript [8, 9].

Many scholarly publications that report original research follow the IMRaD (Introduction, Methods, Results, and Discussion) structure [22–26]. Two other critical sections, the Abstract and the Conclusions, serve as academic summaries that frame this main body of work. This study focuses on the semantic and linguistic quality of these two summary sections. We propose an LLM-based workflow to diagnose two common issues that can detract from the clarity and integrity of these summaries: the inclusion of information not substantiated in the main text and the use of ambiguous pronoun constructs.

Both of these issues undermine the core function of a scholarly manuscript: the effective communication of scientific information. Scientists generally expect that all factual (that is, evidential rather than summative) claims within a summary must originate from and be substantiated by the IMRaD content of the paper [25, 27, 28], where proper context is provided. Introducing new information (an unsubstantiated claim) in a summary section either deprives that information of necessary context or undermines the communicative function of the summary itself. Similarly, ambiguous pronouns (e.g., "it" or a standalone "this") can disrupt narrative flow and complicate comprehension by obscuring the intended antecedent ("The way that many scientists and engineers treat the pronoun it is unsettling but the way that many scientists and engineers treat the word this is criminal") [29]. To address these problems, we developed and tested a series of proof-of-concept (PoC) structured prompts designed to guide an LLM in identifying and flagging these specific issues within a sample text [30]. The prompts and our development process are detailed in the appendices and supporting information.

## 2. Methodology

### 2.1. Materials

The proof-of-concept prompts developed in this study were tested using a single, deliberately selected publication as the input text [30]. The selected test paper was chosen because prior analysis [8, 9] had identified it as containing relevant examples of the textual issues under investigation. The specific sections of the test paper analyzed were the

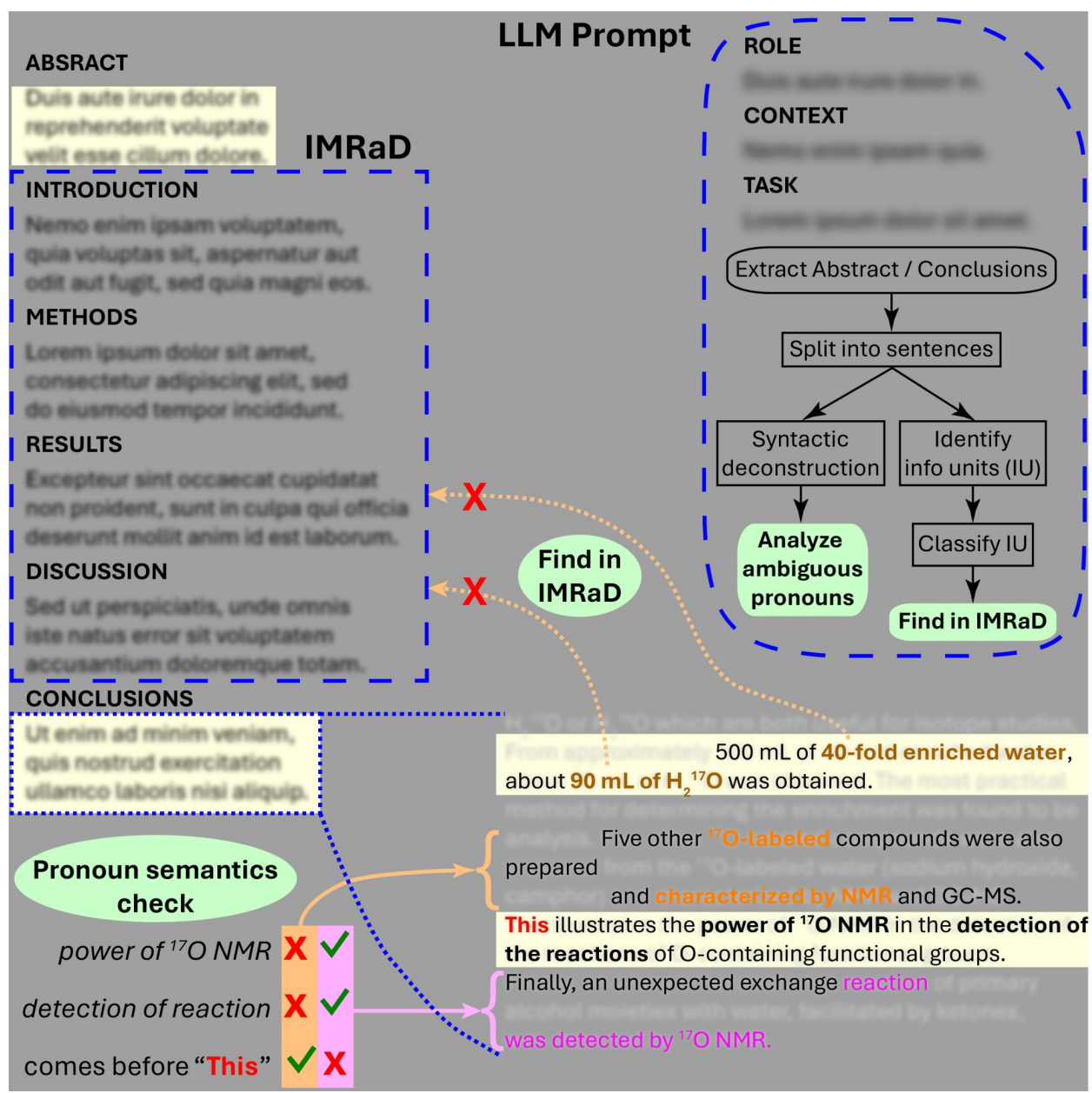

Figure 1. Graphical abstract

Abstract and Conclusions. All prompt development and testing were performed using the Gemini Pro 2.5 Pro model accessed via its official web and mobile interface, as well as via Google AI Studio [5]. Additional testing was also performed using ChatGPT Plus o3 model accessed via its official web interface [6].

## 2.2. Prompt Design

The core prompt engineering strategy employed in this study starts with analysis and hierarchical decomposition of the complex target task, yielding a workflow that seeks to mimic human hierarchical reasoning [31, 32]. Subsequently, the workflow is implemented as modular structured prompts that are designed to elicit hierarchical reasoning when guiding frontier reasoning LLM models. The workflow forms the basis of the top-level **Task** section of a commonly employed role-based design:

- **Role/Persona**: Primes LLM behavior to achieve better prompt adherence and fidelity, particularly for complex domain-specific tasks.
- **Context:** A general task description that may act as a summary or high-level introduction for the complex **Task** section, orienting the model and providing procedural highlights if necessary.
- **Task:** The core hierarchical workflow.
- **Output Format:** General report formatting instructions (step-specific output instructions are commonly embedded within the related block of the **Task** section).
- **Final Instructions:** Core behavior reinforcement.

This modular, structured design offers several benefits. In the present case, the prompts employ a linear workflow that "drills down" through the hierarchical organization of a technical text, starting from the entire input and gradually narrowing focus to "atomic" information units. The core idea is that any intermediate output becomes part of the model's context that can be used in subsequent steps. For example, once the "Conclusions" section is identified and extracted, it becomes a focused piece of the context, and subsequent stages no longer need to address the entire manuscript. This modularity also means that prompt sections can be reused for different prompts involving the same sub-tasks (e.g., identifying a summary section). Equally important is the potential for detailed, structured outputs to provide insights into an LLM's specific failure modes.

These prompts were created through an iterative, three-stage development process:

1. **Manual Task Decomposition:** The analytical goal (e.g., "flag new information") was first broken down into a logical sequence of smaller, well-defined subtasks suitable for an LLM. Initial prompts were drafted for each subtask.
2. **Interactive Testing and Refinement:** The subtask prompts were tested interactively on the test case within a single LLM conversation. This step served to evaluate the viability of the decomposition scheme and refine the clarity of the instructions for each step.
3. **Meta-Prompting for Workflow Integration:** The refined subtask sequence was integrated into a single, comprehensive workflow prompt using meta-prompting techniques - a process where one prompt is used to guide an LLM to develop and refine another, more specialized prompt.

The resulting prompts perform two distinct types of analysis - informational and linguistic - applied separately to the Abstract and Conclusions sections.

## 2.3. Informational Integrity Analysis

This analysis uses a dedicated prompt to execute a multi-phase workflow that involves locating the target section, segmenting it into sentences, and decomposing those sentences into discrete "Information Units" (IUs). Each IU is then classified according to a detailed 13-category schema defined in the prompt. The final phase involves verifying each IU against the IMRaD sections of the paper to flag new or unsubstantiated information.

The use of a custom classification system to guide an LLM's semantic analysis is a conceptual approach explored in our prior work [8]. The 13-category schema employed in this study is a further development of that approach, designed specifically for the nuanced informational content of academic summaries. The schema was developed to be a modular tool applicable to both Abstracts and Conclusions, containing categories common to both summary types as well as categories more specific to one. Each category includes a scope definition, the primary IMRaD section where the information is expected to originate, and verification notes for the LLM. For the present PoC study, the system's primary function was to guide the LLM's verification process by directing its search to the most probable IMRaD source section for any given Information Unit.

## 2.4. Linguistic Clarity Analysis

The workflow for this analysis evolved during development. Our initial approach utilized a basic prompt that relied on the LLM's general semantic capabilities, but early testing revealed this method was not sufficiently robust. This finding motivated the development of a more structured workflow where the LLM performs a highly structured assessment. For each pronoun, the workflow defines a precise "pronoun context" by systematically deconstructing its own clause into its semantic components (e.g., action, subject, concept, and all modifiers). It then performs a component-wise sufficiency check to determine if the pronoun context is properly supported by the context of the identified antecedent.

Judging whether such support is sufficient can be challenging for an LLM, particularly when the pronoun's sentence involves an interpretive "action" (the verb) and abstract concepts (e.g., "This illustrates the power of..."). To account for this scenario, the prompt separates the analysis of the often more abstract interpretive verbs from that of the more concrete "substantive components" of the claim (the object of the verb and its modifiers). The analysis separation is achieved by including a dedicated sufficiency check branch for the "action component". However, it should be noted that potentially abstract "substantive components" (particularly the syntactic head noun of the object) are not treated with similar specificity in the current prompt version.

## 2.5. Testing Protocol

To assess the robustness and consistency of the prompts across two different models, a systematic evaluation was conducted. The prompts were tested against the publication [30] using two different frontier LLMs under two distinct conditions: a "limited context run" (for linguistic analysis) where only the "Conclusions" section was provided as input, and a "full context run" where the entire manuscript was provided. Models were accessed solely through the official web and mobile user interfaces; no API access was employed. All tests were conducted by manual submission of prompt texts. Each prompt was submitted to a new AI chat after completion of the previous one. With ChatGPT, each series was executed within a dedicated project, and the test manuscript was submitted once per series as a project file. With Gemini, the test manuscript was attached to each AI chat from Google Drive. Each experiment was repeated in a series of 20-40 successive runs. Additionally, each series was typically repeated at least twice, with each repetition performed on a separate day. The results of individual outputs were manually collected on spreadsheets for subsequent evaluation of success/failure rates.

## 3. Results

This section presents the results from the evaluation of the two proof-of-concept prompts developed in this study. We first detail the quantitative results from the systematic, multi-run testing of the Linguistic Clarity Analysis prompt. Following this, we report the outcome of the Informational Integrity Analysis prompt when applied to its target test case.

## 3.1. Informational Integrity Analysis

The ground truth for this analysis was established by analyzing the "Conclusions" section of the test paper [30] for information not previously substantiated in the IMRaD sections. The third sentence of the Conclusions states: "From approximately **500 mL of 40-fold enriched water, about 90 mL of $H_2^{17}O$** was obtained." While the input volume of "500 mL" is mentioned in the main text, the specification of "**40-fold enriched water**" (presumably the implied result of the first experimental stage) is not. Furthermore, the output quantity of "**90 mL of $H_2^{17}O$**" is not explicitly stated in the main text, nor can it be directly derived from the data presented in the paper's tables or figures. Therefore, these pieces of information should be flagged as "unsubstantiated claims".

The prompt's performance was evaluated for its ability to identify these two distinct pieces of unsubstantiated information across 20 runs for each model. The number of successful identifications ("hits") for each target is summarized in **Table 1**.

**Table 1.** Successful identifications of unsubstantiated information (out of 20 runs).

|  | "90 mL" | "40-fold" |
|---|---|---|
| ChatGPT Plus o3 | 19 | 0 |
| Gemini Pro 2.5 Pro | 19 | 19 |

## 3.2. Linguistic Clarity Analysis

The ground truth for this analysis hinges on the distinction between a pronoun's intrinsic grammatical ambiguity and its semantic adequacy. While a standalone pronoun like "this" is often considered grammatically vague by default, its semantic adequacy depends entirely on whether its full context is explicitly supported by a clear antecedent. The test

**Table 2.** Performance of the Linguistic Clarity Analysis prompt with ChatGPT Plus o3.

| Series | Context | Runs | Successes | Failures | Success Rate |
|---|---|---|---|---|---|
| B | Limited | 20 | 20 | 0 | 100% |
| A | Full | 19 | 17 | 2 | 90% |
| B | Full | 20 | 16 | 4 | 80% |

Note: One run was excluded from Series A (Full Context) due to the accidental use of an incorrect model version.

**Table 3.** Performance of the Linguistic Clarity AnalysisPrompt with Gemini Pro 2.5 Pro.

| Series | Context | Runs | Successes | Failures | Success Rate |
|---|---|---|---|---|---|
| A | Limited | 21 | 12 | 9 | 55% |
| B | Limited | 40 | 14 | 26 | 35% |
| C | Limited | 40 | 21 | 19 | 55% |
| A | Full | 20 | 14 | 6 | 70% |
| B | Full | 40 | 34 | 6 | 85% |
| C | Full | 40 | 35 | 5 | 90% |

case for this analysis is the second-to-last sentence of the "Conclusions" section in the test case [30]: "This illustrates the power of $^{17}$O NMR in the detection of the reactions of O-containing functional groups." This sentence involves an interpretive claim, and its "pronoun context" can be deconstructed into three main components:

- **Action/verb:** "illustrates"
- **Abstract concept:** "power of 17O NMR"
- **Scope modifier:** "detection of the reactions of O-containing functional groups"

For the reference to be considered semantically adequate, each of these components must be explicitly supported by the antecedent text.

The relevant context in the preceding sentence includes "Five other $^{17}$O-labeled compounds were also prepared ... and characterized by NMR and GC-MS." This antecedent context does not explicitly mention the "detection of any reactions". Furthermore, this passing generic mention of NMR (along with another complementary analytical technique) for routine characterization does not semantically support the claim of illustrating the "power" of the technique. The last sentence *does* mention a reaction detected by NMR and could conceivably act as a formal antecedent in this case, if the last two sentences were swapped (though neither of these sentences is a suitable last sentence in this case), but the sentence following the pronoun cannot serve as a valid antecedent. The pronoun "This", therefore, should be flagged as ambiguous and lacking a clear antecedent in the local context.

The primary success criterion for the quantitative evaluation was the prompt's ability to guide the LLM to correctly identify that the "detection of reactions" component within the target sentence's pronoun context was not explicitly supported by the antecedent text. The performance of the prompt was tested across two LLMs, with test series conducted on different days to assess consistency. Performance was evaluated under both limited context (only the "Conclusions" section provided) and full context (the full manuscript provided) conditions. A total of 201 test runs were conducted with the Gemini Pro 2.5 Pro model, and 59 runs were conducted with the ChatGPT Plus o3 model. The results are summarized in **Tables 2** and **3**.

## 4. Discussion

The results of this PoC study are consistent with the initial hypothesis that structured workflow prompts can guide LLMs to perform specific, non-trivial analytical tasks on the summary sections of scholarly manuscripts. In the Linguistic Clarity Analysis, both models showed high success rates with full context, though their behavior differed in the limited context condition. In the Informational Integrity Analysis, the results suggest potential differences in model capabilities for this specific task.

### 4.1. Interpretation of Findings

The Informational Integrity Analysis (**Table 1**) targeted two unsubstantiated items in the test case. The results suggest a notable difference in how the models handled the verification task. While both models were highly successful at identifying the unsubstantiated quantitative value ("90 mL"), ChatGPT Plus o3 did not identify the unsubstantiated adjectival modifier ("40-fold") in any of the 20 runs (0% success). In contrast, Gemini Pro 2.5 Pro identified both targets with a high success rate (95% for each). A qualitative review of the models' outputs offers insight: Gemini's detailed reasoning for its single failure on the "40-fold" target showed that it correctly identified the phrase but judged it to be a "reasonable summary" - an incorrect but understandable inference. ChatGPT's outputs were consistently more terse, making it difficult to diagnose the precise cause of its failure.

The Linguistic Clarity Analysis (**Tables 2** and **3**) also yielded noteworthy findings. The high success rates achieved with full context by both models (80-90%) are consistent with the robustness of the prompt's architecture. A key observation emerges from the comparison of context conditions. For Gemini Pro 2.5 Pro, limited context was associated with a significant degradation in performance. Conversely, ChatGPT o3 achieved a 100% success rate with limited context, which may indicate a greater ability to operate strictly within the prompt's local constraints for this specific task.

A secondary analysis revealed that the prompt was also effective at guiding the models to identify the unsupported nature of the more abstract "power of NMR" concept. For both models and across all conditions, the success rate for correctly flagging this abstract concept as unsupported was nearly identical to the success rate for flagging the more concrete "detection of reactions" component. This result suggests that the models did not find the abstract nature of the "power" concept to be significantly more challenging than the concrete scope modifier in this particular test case. Finally, a close examination of the outputs revealed that neither model generally produced false positives by flagging other text as problematic.

### 4.2. Limitations

This PoC study has several important limitations. The primary limitation is the use of a single test case for both analyses. While the multi-run, multi-day, and multi-model protocol provides a deep analysis of prompt robustness for these specific instances, the generalizability of the findings is not yet established. The 100% success rate of ChatGPT on the linguistic task, for example, requires testing against a larger corpus to determine if it represents a more fundamental model capability or an artifact of the specific test sentence. Furthermore, the observed temporal instability of Gemini's performance highlights the challenges of using public-facing web interfaces for research.

## Conclusions

The results of this proof-of-concept study are consistent with the initial hypothesis that structured workflow prompts can guide LLMs to perform specific, non-trivial analytical tasks on summary sections of scholarly manuscripts. In the Informational Integrity Analysis, both models demonstrated the same high success rate in identifying an unsubstantiated head of a noun phrase. In contrast, when identifying an unsubstantiated adjectival modifier, Gemini Pro 2.5 Pro model performed equally well, whereas the ChatGPT Plus o3 model completely failed, suggesting a potential difference in how the models handle verification based on a term's syntactic role. For the Linguistic Clarity Analysis, both models performed well with full manuscript context. However, their behavior diverged significantly in a summary-only setting: ChatGPT achieved a perfect success rate, whereas Gemini's performance was substantially degraded. These results suggest that while structured prompting is a powerful and effective methodology for performing targeted analysis of scholarly text, the reliability of the outcome may depend on the interplay between the prompt architecture, the specific model used, and the contextual information provided.

## Acknowledgments

Generative AI use has been an integral part of performed research, including interactive development of prompts via meta-prompting and extensive document revisions. This representative conversational log [33] documents the use of the LLM Gemini (Google) to assist in the iterative revision and refinement of this manuscript. It serves as a demonstration of actively using AI as a peer collaborator during manuscript development.

## Supporting Information

To facilitate direct replication and review of the presented analyses, all materials are available via a view-only link: https://osf.io/nq68y/files/osfstorage?view_only=fe29ffe96a8340329f3ebd660faedd43. This repository includes the *test paper* PDF file (combined manuscript + SI [30]) used as input for analyses and the complete prompt files (in Markdown format, also attached to the manuscript PDF file):

- Informational Integrity Prompt: *ConclusionsClassificationAndReferencesPrompt.md*
- Linguistic Clarity Prompt: *ConclusionsLinguisticAnalysisPrompt.md*

# A. Classification System for Information Units (IU)

> *Note: see PDF attachment file Classification_System_Information_Units_IU.md or [SI](#) for the source Markdown-formatted text.*

This section defines a system of 13 categories for classifying Information Units (IU). These categories describe distinct types of scholarly information that are typically found within, or are suitable for inclusion in, summary sections of an academic manuscript, such as an Abstract or a Conclusions section. All such information is understood to originate from, and be substantiated by, the main IMRaD (Introduction, Methods, Results, Discussion) content of the paper. Each category definition below includes:

- **Scope:** Primarily guides the classification of an Information Unit by defining the nature of the information it contains.
- **Primary IMRaD Location:** Indicates where the detailed, original information is typically first presented in the main paper, guiding where to search for or verify such information.
- **Verification Notes:** Provide criteria for assessing the integrity, appropriate sourcing, and faithful representation of an Information Unit when it appears in, or is being considered for, a summary section (such as an Abstract or a Conclusion). They help ensure that such summary elements accurately reflect the detailed IMRaD content and adhere to good scholarly practice (e.g., not introducing new data in a summary of findings, ensuring limitations are contextually appropriate if mentioned in a summary).

1. **Background, Aim, and Problem Statement:**
   - **Scope:** Information Units that establish brief background/context for the study, AND/OR state the core research question(s), objective(s), hypothesis (hypotheses), or the problem/gap the study was designed to address. (In an Abstract, this often forms the opening statements; in Conclusions, it's typically a focused reminder of the core purpose or problem).
   - **Primary IMRaD Location for First Introduction/Substantiation:** Introduction
   - **Verification Notes:** Verify that this accurately reflects, and does not misrepresent or unduly expand upon, the background, aims, objectives, and problem statement detailed in the Introduction section of the main paper.

2. **Statement of Core Methodology**
   - **Scope:** Information Units concisely describing the primary methods, key experimental design features, main apparatus, population/sample, or principal operational approach used in the study. (Essential for Abstracts; Conclusions would typically only mention methods if using Category 3).
   - **Primary IMRaD Location for First Introduction/Substantiation:** Methods
   - **Verification Notes:** Ensure this is a fair and accurate summary of the main methodologies detailed in the Methods section; it should not introduce methods not mentioned there nor go into excessive detail inappropriate for a summary.

3. **Methodological Highlight (Pivotal Aspect)**
   - **Scope:** Information Units briefly highlighting a novel, critical, or particularly relevant aspect of the study's methodology that was crucial for the results or represents a significant contribution in itself, often emphasizing why it was pivotal or how it impacted the study. (More typical for Conclusions if a methodological point is a key takeaway).
   - **Primary IMRaD Location for First Introduction/Substantiation:** Methods (for full description); Results (for performance/validation data, if applicable).
   - **Verification Notes:** The full description of the highlighted method must exist in the Methods section. If its effectiveness or novelty is part of the highlight, supporting data/evidence should be present in the Results section.

4. **Key Finding / Main Result**

   - **Scope:** Information Units stating a primary outcome, discovery, or observation that directly addresses the study's main aim(s) or research question(s). (Abstracts will present these very concisely).
   - **Primary IMRaD Location for First Introduction/Substantiation:** Results
   - **Verification Notes:** Verify that these are direct statements or accurate summaries of data, figures, tables, or factual statements presented in the Results section. No new data or results should be introduced here that are not in the main paper's Results.

5. **Subsidiary Finding / Secondary Result**

   - **Scope:** Information Units stating a noteworthy outcome or observation not central to the main research question(s) but providing additional insight or supporting main findings. (Rare in Abstracts, more common in Conclusions if space and significance allow).
   - **Primary IMRaD Location for First Introduction/Substantiation:** Results
   - **Verification Notes:** Verify these were previously presented with supporting evidence in the Results section and are not new data points not found in the main paper's Results.

6. **Interpretation of Finding(s)**

   - **Scope:** Information Units explaining the meaning of the study's results, often connecting findings or exploring reasons for outcomes. (In an Abstract, interpretations are typically very concise and tied directly to key findings; Conclusions may offer slightly more elaborated summaries of interpretations).
   - **Primary IMRaD Location for First Introduction/Substantiation:** Discussion
   - **Verification Notes:** Ensure interpretations are consistent with, and (if in Conclusions) concisely summarize, more detailed interpretations in the Discussion, grounded in Results. Abstracts will offer very brief interpretations.

7. **Answer to Research Question / Resolution of Hypothesis**

   - **Scope:** Information Units explicitly stating how the study's findings answer the initial research question(s) or confirm/refute/modify the initial hypothesis (hypotheses). (In an Abstract, this is often a direct and concise statement).
   - **Primary IMRaD Location for First Introduction/Substantiation:** Discussion
   - **Verification Notes:** Verify this answer/resolution concisely reflects detailed arguments and evidence from the Discussion, which links to Results. Ensure no new claims are made beyond this.

8. **Comparison with Existing Literature / Contextualization**

   - **Scope:** Information Units relating the study's findings to existing knowledge, theories, or previous research, noting consistencies, contradictions, or extensions. (Generally rare and very brief in Abstracts; more common in Conclusions as a summary of key comparisons from the Discussion).
   - **Primary IMRaD Location for First Introduction/Substantiation:** Discussion (primarily for detailed comparisons); Introduction (for foundational context).
   - **Verification Notes:** Confirm statements summarize comparisons and contextualization already explored in the Discussion. Abstracts rarely contain this.

9. **Statement of Broader Significance / Impact**

   - **Scope:** Information Units articulating the wider importance, contribution, or potential value of the study's findings to its specific field or to society more generally. (A key component for both Abstracts and Conclusions).
   - **Primary IMRaD Location for First Introduction/Substantiation:** Discussion
   - **Verification Notes:** Check that these statements are logical extensions of findings and interpretations, with the detailed arguments supporting this significance in the Discussion.

10. **Practical Application / Recommendation**

    - **Scope:** Information Units suggesting how the study's findings could be translated into real-world applications, or making specific recommendations for practice, policy, design, or intervention. (May be very concise in Abstracts; more elaborated in Conclusions if based on Discussion).
    - **Primary IMRaD Location for First Introduction/Substantiation:** Discussion
    - **Verification Notes:** Ensure these stem from findings and interpretations explored and justified in the Discussion.

11. **Acknowledgement of Study Limitation(s)**

    - **Scope:** Information Units identifying constraints, weaknesses, caveats, or boundaries related to the study's design, methodology, sample, or the generalizability of its findings. (Very rare in Abstracts; more common and important in Conclusions for balance).
    - **Primary IMRaD Location for First Introduction/Substantiation:** Discussion (most common); Methods (for purely methodological limitations).
    - **Verification Notes:** Verify these are consistent with limitations detailed in the Discussion or Methods. No new, unmentioned limitations should appear in Conclusions. Abstracts typically omit these.

12. **Suggestion for Future Research / Outlook**

    - **Scope:** Information Units proposing specific directions for subsequent studies, new research questions arising from the current findings, or areas needing further investigation, including a broader outlook. (May be very brief or absent in Abstracts; more common and detailed in Conclusions).
    - **Primary IMRaD Location for First Introduction/Substantiation:** Discussion
    - **Verification Notes:** Check that these suggestions logically arise from the study and summarize detailed suggestions from the Discussion. Abstracts rarely detail this.

13. **Overall Concluding Remark / Take-Home Message**

    - **Scope:** An Information Unit (IU) (often a full sentence when not further chunked, or a dominant clause) providing a final, high-level summary that encapsulates the main essence of the study's findings and their importance. (This is the culminating statement for both Abstracts and Conclusions).
    - **Primary IMRaD Location for First Introduction/Substantiation:** Content derived from Results and Discussion; specific phrasing is unique to summary sections.
    - **Verification Notes:** The message must be a fair and accurate representation of substantiated contributions detailed in Results/Discussion and should not introduce new substantive claims.

## B. Fair Use Statement - Sharing Test Paper

1. **Identification of Copyrighted Material:**

   - **Work:** "Enrichment of H217O from Tap Water, Characterization of the Enriched Water, and Properties of Several 17O-Labeled Compounds".
   - **Authors:** Brinda Prasad, Andrew R. Lewis, and Erika Plettner.
   - **Publication:** *Anal. Chem.* 2011, 83, 1, 231-239.
   - **DOI:** 10.1021/ac1022887.
   - **Publisher/Copyright Holder**: American Chemical Society.
   - **Material Shared:** A combined digital file containing the full text of the aforementioned article and its complete associated Supporting Information (SI).

2. **Sharing Mode:**

   - **Resource:** Private Open Science Framework (OSF) project repository.
   - **Location:** https://osf.io/nq68y/files/osfstorage?view_only=fe29ffe96a8340329f3ebd660faedd43.
   - **Protection Measures:** Due to private nature, the resource should not be indexed by search engines.

3. **Assertion of Fair Use:**

   The sharing of this copyrighted material is undertaken for specific, limited purposes, believed in good faith to constitute "fair use" under Section 107 of the U.S. Copyright Act (or applicable analogous principles in other jurisdictions).

4. **Purpose and Character of Use (Factor 1):**

   - **Non-Profit Educational and Research:** The use is strictly for non-commercial research and educational purposes, specifically within the context of scholarly critique and the advancement of research methodology.
   - **Transformative Use:** The work is not merely being reproduced; it is fundamentally repurposed as a research specimen for critical analysis. Its primary function in this context is not to convey its original purported findings, but to serve as the subject of rigorous evaluation and methodological demonstration.
   - **Critique and Commentary:** A core purpose is to conduct and disseminate a detailed, peer-review-like critique of the article's methodology, analysis, and conclusions. This critique identifies significant flaws within the original work.
   - **Advancement of Knowledge & Methodology:** The use includes the development and demonstration of a novel AI-driven prompt/technique for manuscript analysis. Sharing the specimen (the article + SI file) is integral to demonstrating and enabling the verification and further development of this new analytical method.

5. **Nature of the Copyrighted Work (Factor 2):**

   - The original work is a published scholarly article, typically factual in nature, a category often amenable to fair use for purposes of scholarship and critique.
   - However, the conducted analysis (central to this project) has revealed substantial flaws impacting the reliability and validity of the work's core research findings as presented. This impacts the assessment of its nature in the context of this specific use.

6. **Amount and Substantiality of the Portion Used (Factor 3):**

   The entire article and its complete Supporting Information are utilized and shared in a combined format.

   Justification: This amount is essential and necessary for the stated purpose. A comprehensive critique, akin to thorough peer review or forensic analysis, requires examination of the whole work, including all data and methods presented in the SI. Evaluating the integrity and validity of the research necessitates access to the complete context. Furthermore, the development and validation of the AI analysis prompt require the complete text as input. The combined file format, not available directly from the publisher, was the specific subject of the analysis.

7. **Effect of the Use upon the Potential Market for or Value of the Copyrighted Work (Factor 4):**

   - **No Harm to Legitimate Market:** This use is not intended to, nor is it likely to, negatively impact the legitimate market or value of the original copyrighted work. The publisher's market relies on the perceived value of the article as a source of valid scientific findings.
   - **Critique Reveals Lack of Value:** The critique resulting from this research demonstrates fundamental flaws undermining the article's claimed scientific value. Therefore, sharing the work specifically in this context (as a specimen for critique and methodological development) does not substitute for or usurp the market for the work based on its originally purported merits, as those merits are shown to be compromised. Dissemination for critique serves the public interest by highlighting these issues, distinct from fulfilling the original market demand.
   - **Controlled Access for Verification via Private OSF Project:** To ensure transparency and enable independent verification and follow-on research by interested parties engaging with the publicly disseminated research critique manuscript [TBD], the combined article + SI file (serving as the direct supporting evidence and test specimen) is hosted within a private Open Science Framework (OSF) project. A view-only link to this private project will be provided alongside the manuscript.
   - **Minimized Risk of Unintended Use:** This method ensures that access is granted specifically to individuals who are actively reviewing or assessing the research critique presented in the manuscript. The private nature of the OSF project prevents general public discovery through search engines, and the view-only restriction prevents facile downloading and redistribution. Access requires the specific link obtained from the context of the critique manuscript.
   - **Purpose Remains Transformative, Not Substitutive:** By utilizing a controlled-access, view-only repository linked directly to the research critique, this approach provides the necessary transparency for verification while strictly limiting potential downstream use and eliminating broad public access. This method strongly reinforces that the purpose is critique and verification (transformative uses), not market substitution for the original work's questioned scientific claims, thereby minimizing any potential harm to a legitimate market.

8. **Conclusion:**

   Based on the non-profit, educational, highly transformative nature of the use (critique, commentary, methodological advancement), the necessity of using the entire work for these specific purposes, and the argument that this use does not harm the legitimate market value due to the work's identified flaws and the distinct purpose of sharing, this distribution is asserted to be fair use.

   This material is intended solely for the recipient(s) for purposes directly related to verifying, understanding, or building upon the presented critique and methodological research. Further distribution is not permitted. Copyright remains with the original holder(s).